\documentclass[10pt,twocolumn,letterpaper]{article}

\usepackage{iccv}
\usepackage{times}
\usepackage{epsfig}
\usepackage{graphicx}
\usepackage{amsmath}
\usepackage{amssymb}
\usepackage{times}
\usepackage{epsfig}
\usepackage{graphicx}
\usepackage{amsmath}
\allowdisplaybreaks
\usepackage{amssymb}
\usepackage{graphicx}
\usepackage{booktabs}
\usepackage{textcomp}
\usepackage{dsfont}
\usepackage{color}
\usepackage{subcaption}
\usepackage{xfrac}
\usepackage{xcolor}
\usepackage{dblfloatfix}
\usepackage{balance}
\usepackage{float}
\usepackage[ruled,noend,noline]{algorithm2e}

\usepackage[breaklinks=true,bookmarks=false]{hyperref}

\iccvfinalcopy 

\def\iccvPaperID{9234} 

\ificcvfinal\fi

\begin{document}

\title{Hierarchical Road Topology Learning for Urban Map-less Driving}

\author{Li Zhang \quad Faezeh Tafazzoli \quad Gunther Krehl \quad Runsheng Xu \quad Timo Rehfeld\\ Manuel Schier \quad Arunava Seal\\
Mercedes-Benz R\&D NA\\
\tt\small \{li.lz.zhang, faezeh.tafazzoli, gunther.krehl, runsheng.xu, timo.rehfeld,\\ 
\tt\small manuel.schier, arunava.seal\}@daimler.com
}
\maketitle
\ificcvfinal\thispagestyle{empty}\fi

\begin{abstract}
     The majority of current approaches in autonomous driving rely on High-Definition (HD) maps which detail the road geometry and surrounding area. Yet, this reliance is one of the obstacles to mass deployment of autonomous vehicles due to poor scalability of such prior maps. In this paper, we tackle the problem of online road map extraction via leveraging the sensory system aboard the vehicle itself. To this end, we design a structured model where a graph representation of the road network is generated in a hierarchical fashion within a fully convolutional network. The method is able to handle complex road topology and does not require a user in the loop.
\end{abstract}

\section{Introduction}

\label{sect:introduction}
Autonomous vehicles tend to rely on data-hungry perception algorithms to comprehend their surroundings. They are loaded with a constellation of sensors collecting data from the ambient environment, such as position and dimensions of the surrounding objects, weather condition, and traffic, but also large, detailed, and accurate global maps.\\
Maps are an indispensable component of self-driving technology. The unique needs of autonomous vehicles necessitate a new class of HD maps for prior map-based localization, modeling the road surface at centimeter-level accuracy, enabling an autonomous vehicle to confidently deduce its position with respect to the ambient environment. With such strong prior knowledge, an autonomous vehicle is able to enhance its perception and react better to the events on the road beyond the reach of on-board sensors, facilitating its interactions with other traffic participants. As such, HD maps, generally provide 3D geometric and semantic information on static and physical parts of the world, including lane boundaries, intersections, crosswalks, parking spots, stop signs, and traffic lights. These static maps are computed offline, typically using the sensors of the self-driving vehicle itself, although manual annotations or modifications are often required.
\begin{figure}
     \centering
     \begin{subfigure}[b]{0.23\textwidth}
         \centering
         \includegraphics[width=0.98\textwidth]{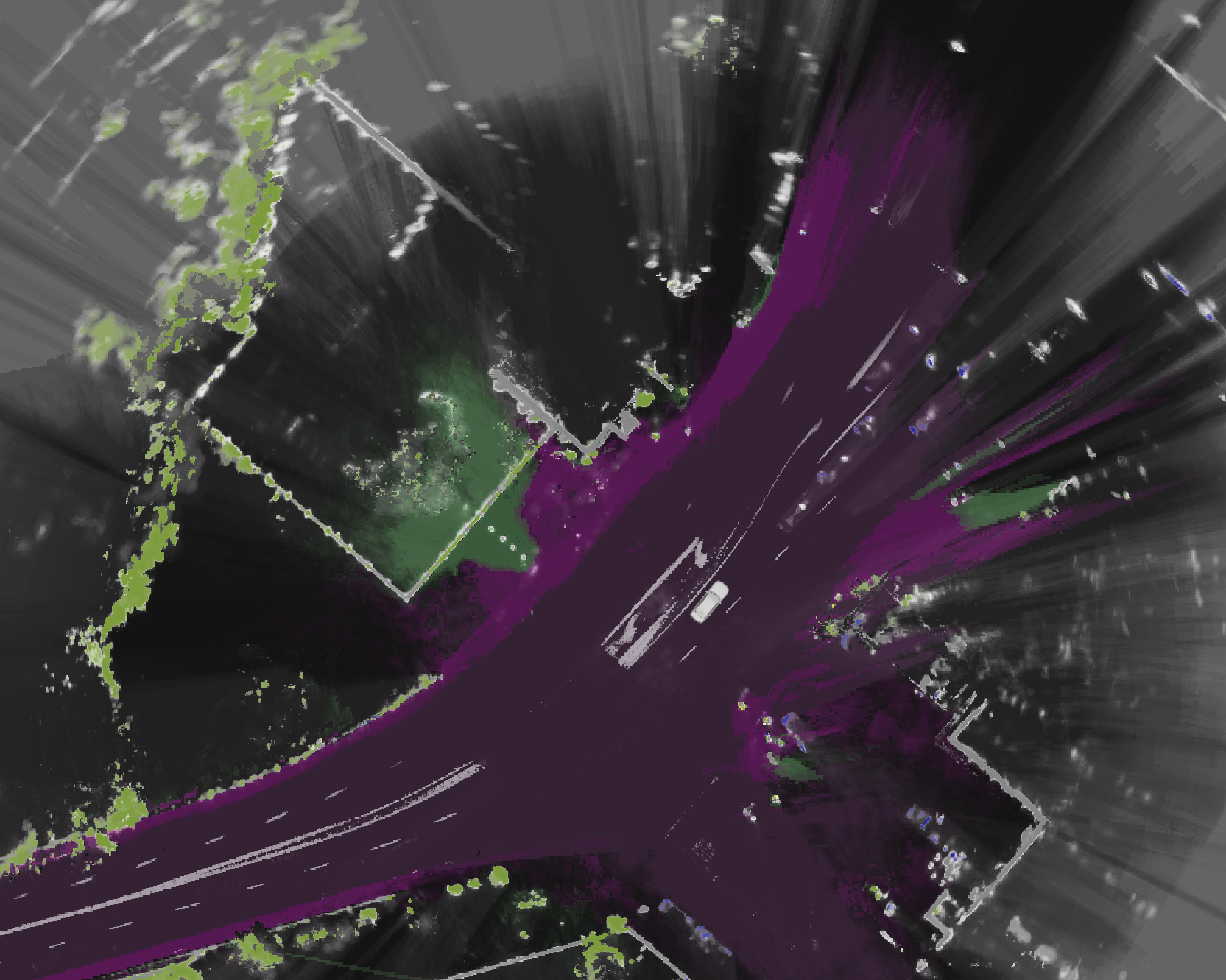}
         \caption{Input grid map}
         \label{fig:inputlayers_overlayed}
     \end{subfigure}
     \begin{subfigure}[b]{0.23\textwidth}
         \centering
         \includegraphics[width=0.98\textwidth]{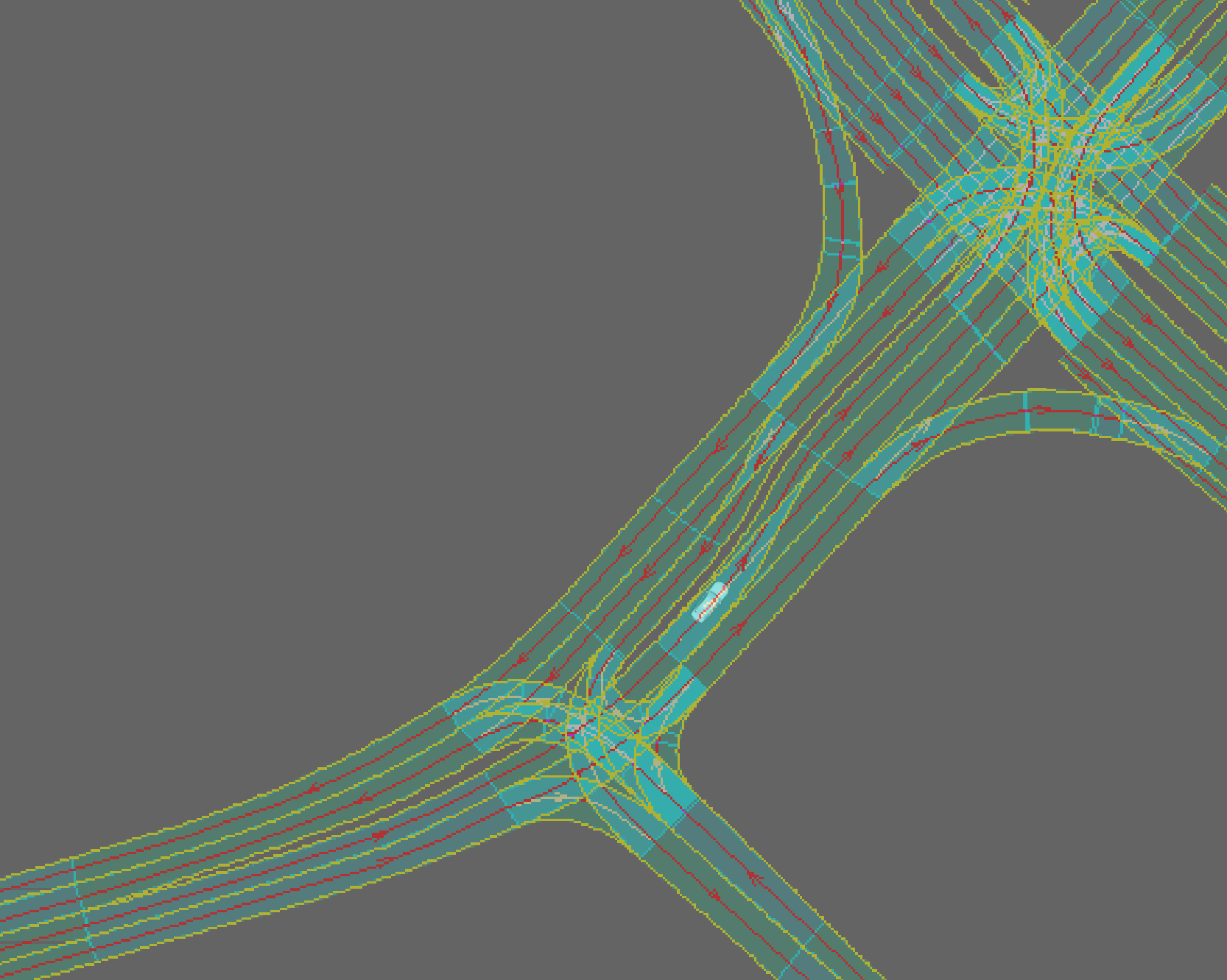}
         \caption{HD map}
         \label{fig:hdmap}
     \end{subfigure}
     \begin{subfigure}[b]{0.23\textwidth}
         \centering
         \includegraphics[width=0.98\textwidth]{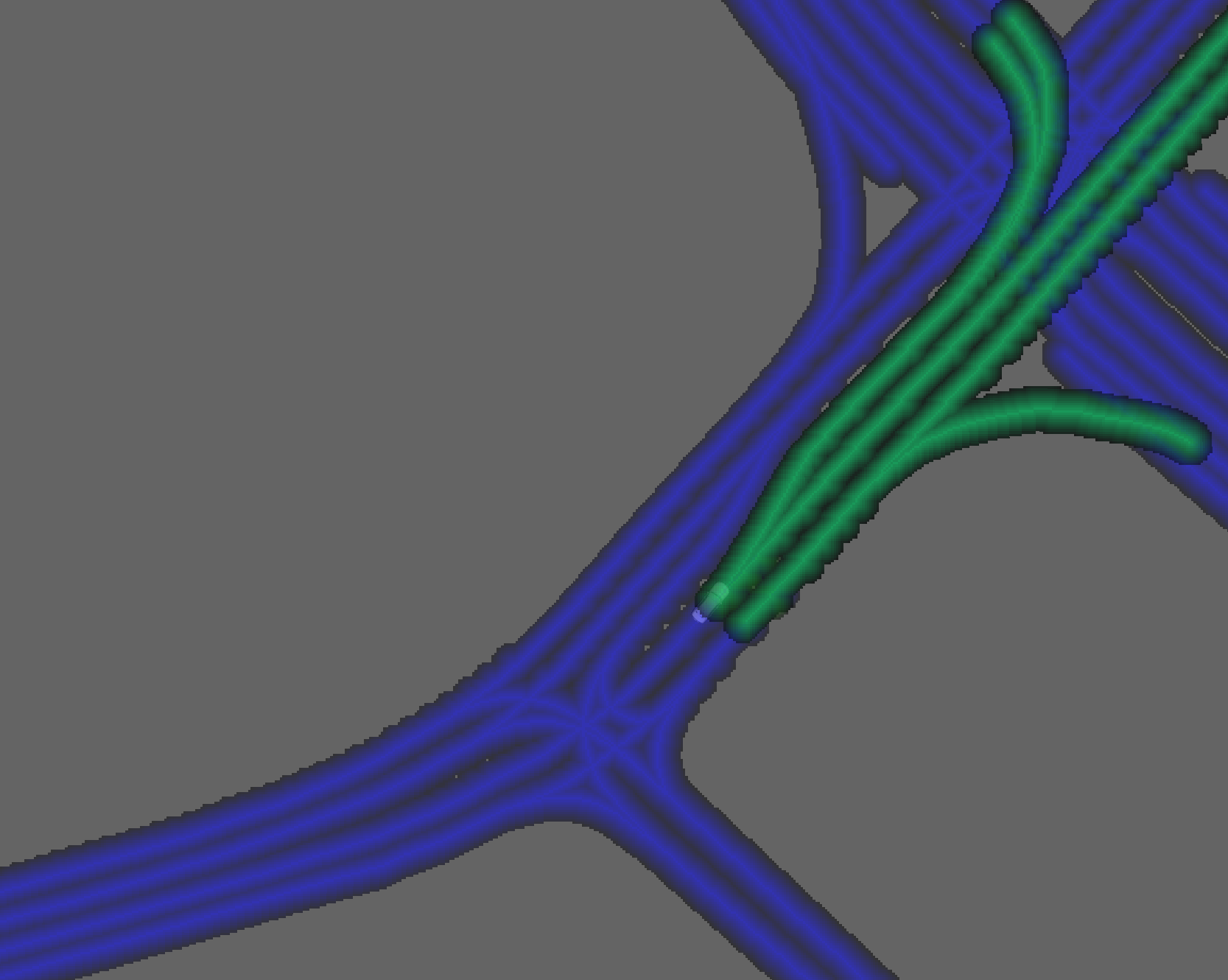}
         \caption{Ego drivable lanes}
         \label{fig:fagulispace}
     \end{subfigure}
     \begin{subfigure}[b]{0.23\textwidth}
         \centering
         \includegraphics[width=0.98\textwidth]{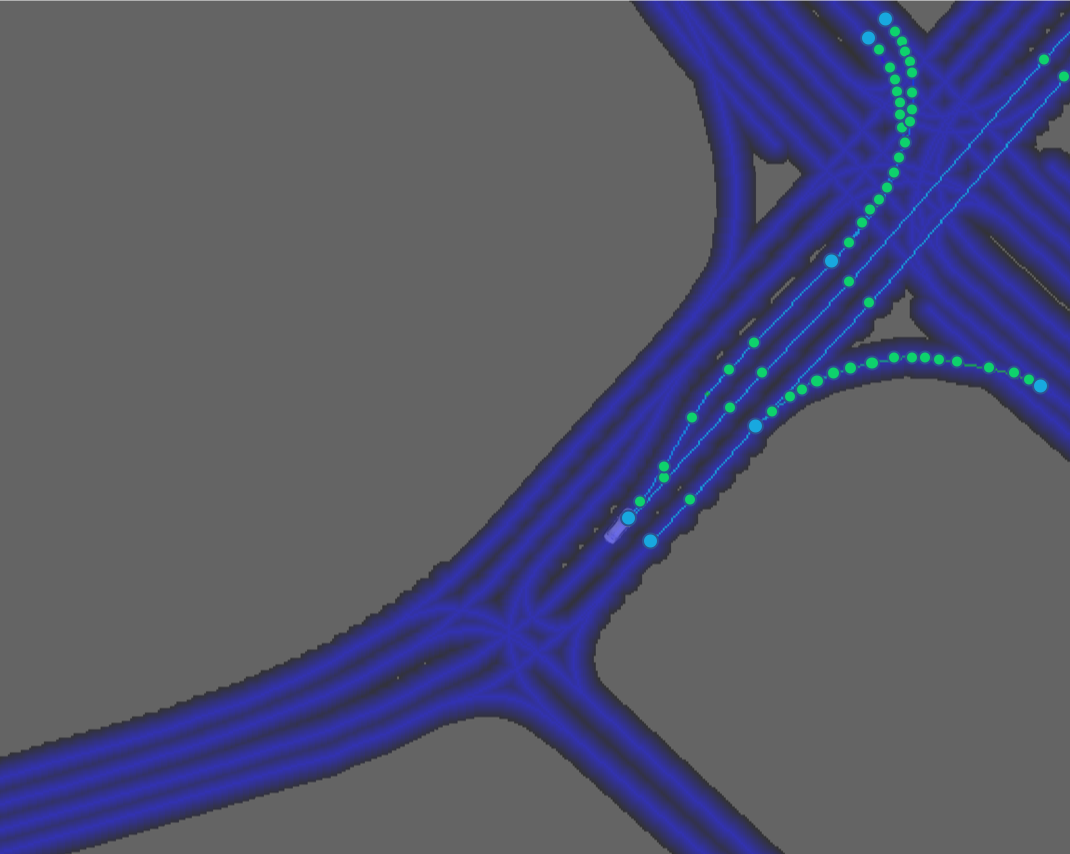}
         \caption{Road topology}
         \label{fig:roadtopology}
     \end{subfigure}
     \caption{Example of autonomous vehicle\textquotesingle s potential drivable lanes and road topology.}
     \label{fig:drivabellanes}
\end{figure}

While this paradigm has served to facilitate Autonomous Driving, such dependence on detailed prior maps is undesirable for global scale, as it requires a wealth of information about the geometry and traffic rules of every single road, and consequently, large volumes of data and storage space. Such maps are not only expensive to store on-board autonomous vehicles, but also very laborious to create and maintain \cite{homayounfar2018hierarchical}. Furthermore, it is of extreme importance for the maps to reflect the latest state and components of roads, e.g. repainted road markings, blocked roads, construction sites, at all times. This further compounds the problem and renders this paradigm impractical at scale. 
\begin{figure*}[tb!]
     \centering
     \includegraphics[width=.94\linewidth]{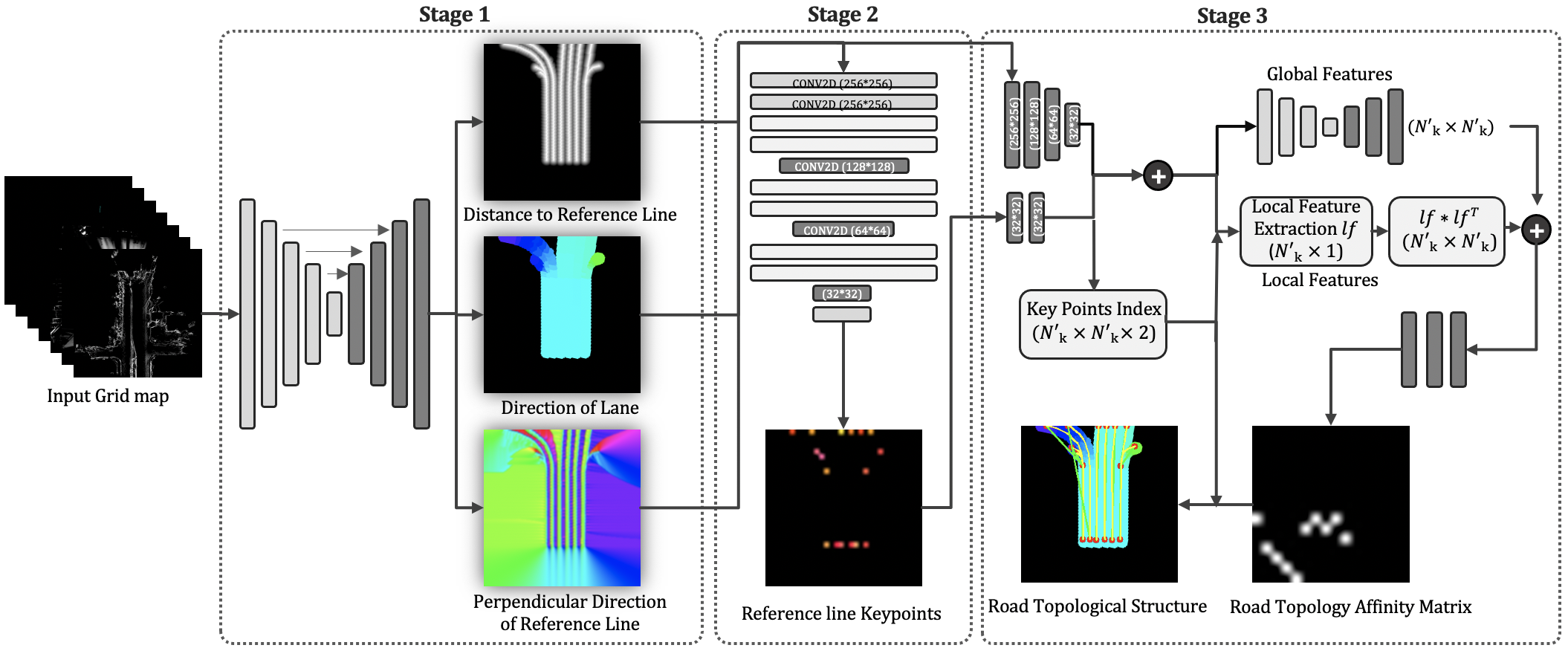}
     \caption{Overview of our road topology learning pipeline. Our model is a multi-stage, multi-task network trained to sequentially create the map components starting from the segmented lanes up to the complete road topology.}
     \label{fig:fusionmap_pipeline}
\end{figure*}

To enhance the practicality and scalability of a self-driving vehicle, we herein propose a solution to transition from heavily relying on HD maps to adopting a methodology with online mapping features. This solution, thereby, eliminates the need for the creation, manipulation, and maintenance of highly accurate maps. The system, as a result, is rendered more agile by adapting to road conditions and does not require precise localization. Figure \ref{fig:fusionmap_pipeline} depicts an overview of our proposed method.\\
We devise a learning methodology where the road condition and its components are learned in its current state, independent of the road complexity and number of lanes. The model receives snapshots of the vehicle\textquotesingle s surroundings reflecting instantaneous environmental condition as well as the road structure and obstructions. Then, it predicts the road topology in a hierarchical fashion; detecting the ego vehicle\textquotesingle s drivable lanes at the low level and, subsequently, connecting this information to a global topological map for robust navigation. In the context of maps, the method produces a road network map, i.e., graph where edges are polylines corresponding to road segments, and vertices represent spatial coordinates of start, end, and fork points of each lane segment. This map dynamically varies with respect to ego vehicle\textquotesingle s position and direction to contain only the relevant information for vehicle\textquotesingle s planning (Figure \ref{fig:drivabellanes}).

\section{Related Work}

\label{sect:background}
To reduce the dependency on maps, several techniques have been developed focusing mainly on predicting drivable routes, which can then be used for generating path proposals. Traditional methods establish connectivity by incorporating contextual priors such as color and texture information \cite{kong2010general, chiu2005lane} or road geometry \cite{dickmanns1992recursive, kuhnl2012spatial}. Some approaches leverage the environment structure and rely on distinct features, such as lane markings and curbs \cite{li2016deep, topfer2014efficient, neven2018towards, beck2014non, lee2018development}. To extend these systems to complex urban environments and rural or undeveloped areas in the absence of clear or consistent lane markings, a class of approaches cast the problem as semantic segmentation to capture large spatial context \cite{suleymanov2018inferring, liang2019convolutional, lee2017vpgnet, he2016accurate, barnes2017find}, or estimate the lane geometry as well as the semantics of each lane \cite{meyer2018deep, homayounfar2018hierarchical}.

To enable navigation and drive in the absence of detailed maps based on a comprehensive understanding of the immediate environment while following simple higher level directions, some approaches include rough map priors as a baseline. In \cite{ort2019maplite}, a map-less driving framework is proposed that combines topometric maps with a LiDAR-based perception system for local navigation. The global topological localization and the corresponding graph search are performed based on the open street map (OSM) data. A decision-making architecture is proposed in \cite{artunedo2019decision} that obtains a global route from OSM and generates driving corridors, which are then adapted and bounded using a vision-based lane detection algorithm and a probabilistic grid-based corridor reduction. The drawback of such solutions, however, is that they cannot reason about roads absent in the initial coarse map.

A class of methods focus on road mapping from aerial images, in which pixel-level segmentation is typically combined with graph-based optimization \cite{mnih2010learning, mattyus2017deeproadmapper, marmanis2016semantic}. Such approaches, however, are usually adopted to merely provide local information about the presence of roads. The detailed information including the inter-connectivity of road segments is provided later through an error-prone post-processing stage. To eliminate such intermediate representation, some methods expand the road tree based on certain footprints or produce the road network directly from a CNN \cite{bastani2018roadtracer}. Although such road topologies are very useful for routing purposes, in the context of autonomous driving, they do not provide the level of detail and accuracy required for safe localization and planning.

In the domain of offline mapping, instead of modeling the geometry of each lane, \cite{homayounfar2019dagmapper} proposes to parameterize the unknown lane graph as a Directed Acyclic Graphical model (DAG) and predicts structured outputs such as a polyline. In such approaches, the data is usually collected by driving each path multiple times, and hence provides much denser information compared to online approaches. An online lane detection is proposed in \cite{homayounfar2018hierarchical} which predicts a structured representation of lane boundaries in the form of polylines. The method exploits a hierarchical Recurrent Neural Network (RNN) to extract them from top-down LiDAR point cloud, where one RNN decides on adding new lanes, while the second RNN predicts the vertices along the lane. The experiments, however, focus mainly on simple topologies in highway scenarios.

\section{Hierarchical Road Topology Learning}
To facilitate map-less autonomous driving, we herein propose a hierarchical map-learning methodology, which does not suffer from the dependency on HD maps and enables the representation of road topology purely based on the sensory system aboard the vehicle. In this methodology, a road topology is defined as a set of keypoints and their relative connectivity, each of which representing a lane segment. As demonstrated in Figure \ref{fig:fusionmap_pipeline}, our model is a multi-stage, multi-task network trained to sequentially create the map components starting from the drivable lanes up to the complete road topology.
\subsection{Input Parameterization}
\label{sect:data}
\begin{figure}
     \centering
     \begin{subfigure}[b]{0.49\linewidth}
         \centering
         \includegraphics[width=0.85\linewidth]{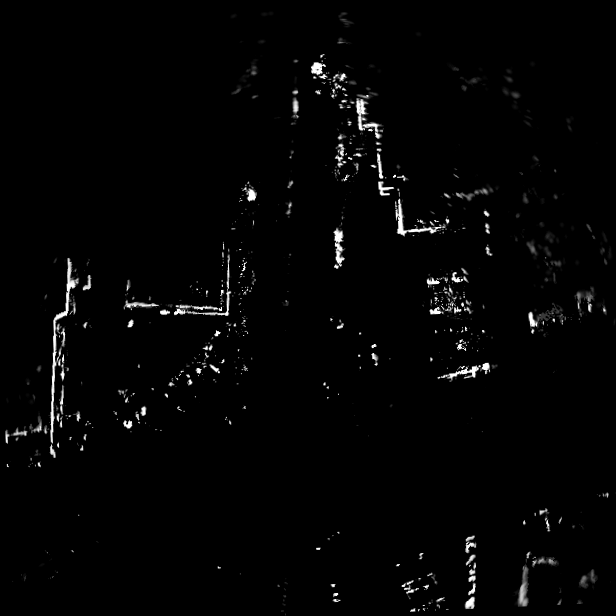}
         \caption{Occupancy}
         \label{fig:inputlayers_occupancy}
     \end{subfigure}
     \begin{subfigure}[b]{0.49\linewidth}
         \centering
         \includegraphics[width=0.85\linewidth]{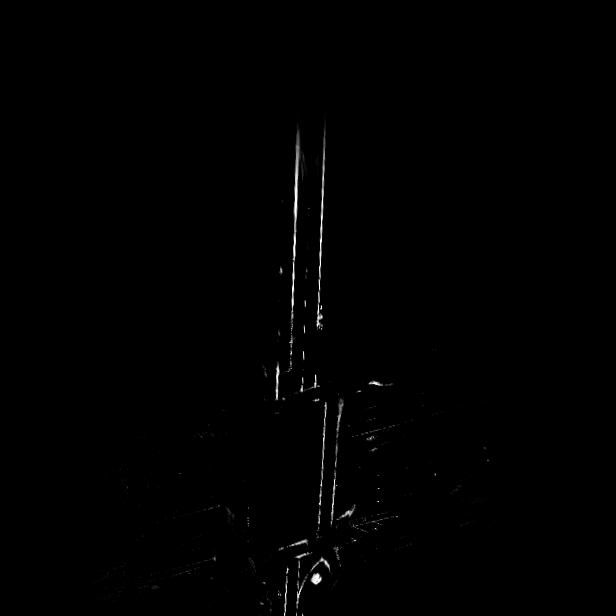}
         \caption{Ground markings}
         \label{fig:inputlayers_groundmarkings}
     \end{subfigure}
     \begin{subfigure}[b]{0.49\linewidth}
         \centering
         \includegraphics[width=0.85\linewidth]{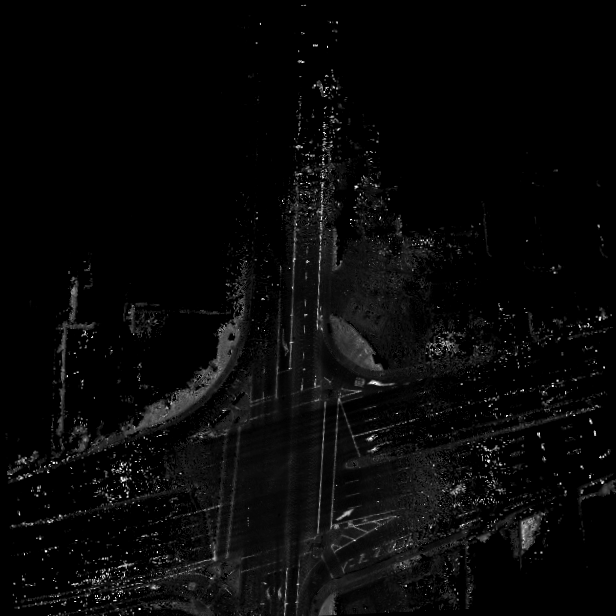}
         \caption{LiDAR intensity}
         \label{fig:inputlayers_intensity}
     \end{subfigure}
     \begin{subfigure}[b]{0.49\linewidth}
         \centering
         \includegraphics[width=0.85\linewidth]{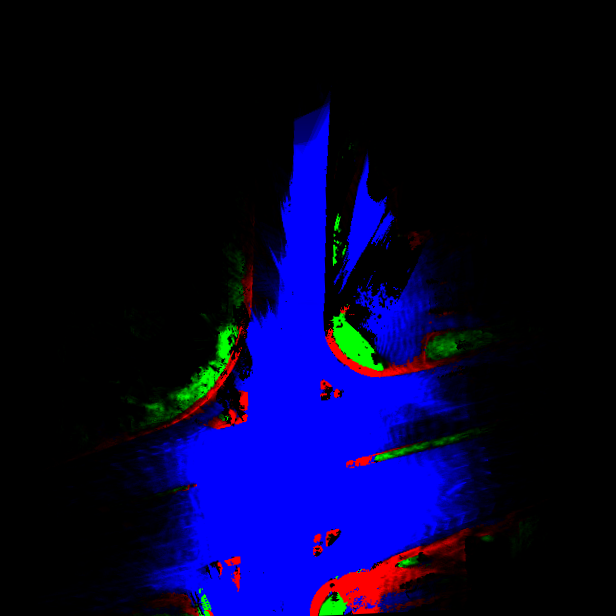}
         \caption{Ground semantics}
         \label{fig:inputlayers_groundsemantics}
     \end{subfigure}
        \caption{Input image channels. For \ref{fig:inputlayers_occupancy}, \ref{fig:inputlayers_groundmarkings}, and \ref{fig:inputlayers_intensity}, higher brightness corresponds to higher existence probability. The data in \ref{fig:inputlayers_groundsemantics} is color-coded to differentiate drivable road, sidewalk, and terrain.}
        \label{fig:inputlayers}
\end{figure}

We encode the environment around the ego vehicle as rasterized bird\textquotesingle s eye view image that contains multiple channels of information: occupancy, ground semantics, ground markings from camera, and ground intensity from LiDAR. The occupancy channel combines measurements from LiDAR and camera that have been semantically classified as obstacle, using Dempster-Shafer rule of combination~\cite{Dempster1968}. The ground semantics channel accumulates information about the location of drivable road, sidewalk, and terrain. The semantic information used to differentiate ground from obstacle measurements is extracted using camera and LiDAR data in a prior step, following \cite{HernandezJuarez2019SlantedSA} and \cite{piewak2018improved}. The model of \cite{HernandezJuarez2019SlantedSA} is further extended to also infer road markings in the camera image, which are accumulated in the ground markings channel. For all channels, measurements are accumulated not only across sensors, but also over time (using ego-motion correction), to have a single holistic and temporally stable input representation. The result is an image
$I \in \mathbb{R}^{H \times W \times C}$, where $C$ is the number of channels. In our experiments, we use an encoding where the vehicle is always positioned at the bottom \sfrac{1}{4} of the image. Figure \ref{fig:inputlayers} shows an example of the channels employed in our proposed approach.
\subsection{Scope Definition}
\label{sect:scope_definition}
Following the intuitive understanding of human\textquotesingle s driving horizon, the definition of road topology is limited to the ego vehicle\textquotesingle s perception range. This definition applies to all lane segments topologically connected to the the lane segment ego vehicle is currently in, and are reachable by moving forward, right, or left. The topology gets updated as the vehicle moves and receives new measurements. To simulate such behavior in the training data received from HD maps, an orientation constraint is applied where each lane segment should also have an acute angle with the ego vehicle\textquotesingle s yaw. Figure \ref{fig:drivabellanes} depicts the above potential drivable space in green overlayed on all the possible routes extracted from HD map depicted in blue. The corresponding road topology defined for the ego vehicle\textquotesingle s potential drivable lanes is displayed in Figure \ref{fig:roadtopology}.
\subsection{Stage 1: Feature Extraction}
\label{sect:drivable_lane_estimation}
The most basic information required for driving is the potential drivable lanes for the autonomous vehicle. The first stage of our architecture, hence, outlines the rough sketch of such areas in $I$, following the scope defined in \ref{sect:scope_definition}. We adopt an encoder-decoder architecture \cite{chaurasia2017linknet}, to aggregate multi-scale features and also preserve spatial information at each resolution. 

The network outputs three representations of the drivable lanes with the same spatial resolution as $I$. The location of the lanes are encoded as a truncated inverse distance transform image $R \in \mathbb{R}^{H \times W \times 1}$ that labels each pixel in $I$ with its relative distance to the closest reference line. To simplify the lane representation and customize the output for future planning purposes, reference line is chosen as the center line of a lane segment in the HD map. In contrast to predicting binary outputs at the lane level, the inverse distance transform of reference line encodes more information about the ideal location of the ego vehicle with respect to the road boundaries. The direction of each lane is represented as $D \in \mathbb{R}^{H \times W \times 3}$, an HSV color-encoded image with continuous values where each pixel in the potential drivable lanes reflects the orientation of the closest reference line. Lastly, the network predicts the perpendicular direction map $P \in \mathbb{R}^{H \times W \times 3}$, encoding the normal directions to the closest reference line. These features are exploited in the later stages of the network to contribute to the hierarchical definition of the map towards road topology prediction (Figure \ref{fig:fusionmap_pipeline}). 

The parameters of the first-stage model are optimized by minimizing a weighted combination of the reference line detection loss $l_{R}$ and the direction estimation losses $l_{D}$ and $l_{P}$:
\begin{align}
  l_{stage1} (I) & = l_{R} (I) + \lambda_{1}l_{D} (I) + \lambda_{2}l_{P} (I) \; 
\end{align}
Both the inverse distance transform and direction map estimation tasks are treated as regression. All three losses are defined as the sum of cosine similarity and L1.
\subsection{Stage 2: Keypoints Generation}
\label{sect:road_graph_generation}
\begin{figure}
     \centering
     \begin{subfigure}[b]{0.97\linewidth}
         \centering
         \includegraphics[width=0.8\linewidth]{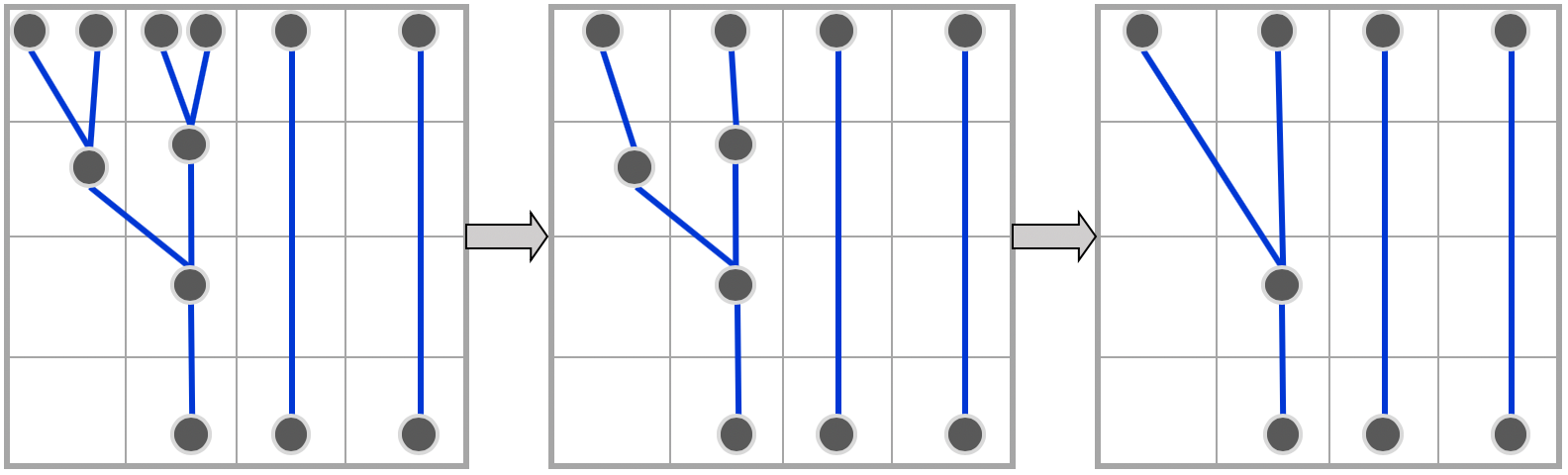}
         \caption{Input graph topology pruning process}
         \label{fig:keypointsgrid_pruning}
     \end{subfigure}
     \begin{subfigure}[b]{0.48\linewidth}
         \centering
         \includegraphics[width=0.58\linewidth]{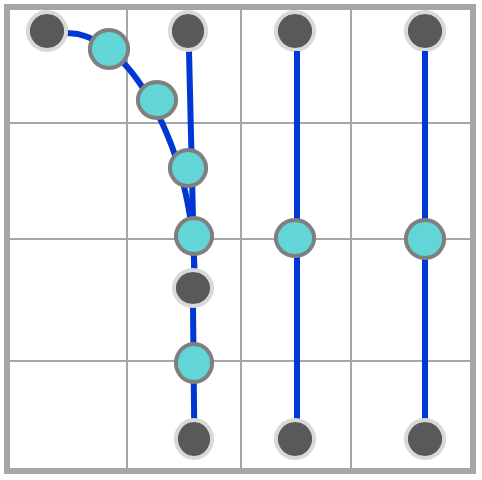}
         \caption{HD map points}
         \label{fig:refline_points_map}
     \end{subfigure}
     \begin{subfigure}[b]{0.48\linewidth}
         \centering
         \includegraphics[width=0.58\linewidth]{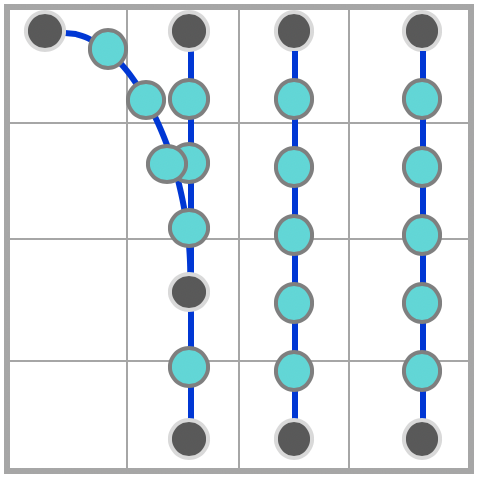}
         \caption{Interpolated points}
         \label{fig:refline_points_gt}
     \end{subfigure}
     \caption{Reference line points definition.}
     \label{fig:refline_gt_preparation}
\end{figure}

Having an estimate of ego vehicle\textquotesingle s drivable lanes and their corresponding features, the second stage serves as an approximation to the baseline of a graph representation of the road by predicting the graph nodes referred to as keypoints , $p(K | R, D, P)$, where $K$ represents the corresponding keypoint grid. Towards this goal, a topology graph is characterized by a set of nodes and their connections.\\
The graph generated based on the driving horizon defined in \ref{sect:scope_definition} might still be very complex, for example, based on the grid map resolution some nodes might be located very close to each other. In the snapshot understanding of road topology, such level of complexity is not required and can be slightly simplified. Hence, to facilitate the learning process, considering the fact that the standard lane width in the United States is $\sim 3.65\:m$, the resolution of keypoint grid is set to $32\times 32$ pixel ($2.08\:m/pixel$), where each pixel represents a $8\times 8$ area of the original perceptive field. To optimize the road graph and reduce the number of parameters and assure that the keypoints in neighboring lanes do not fall into the same grid cell, only one keypoint is kept per cell, which can be either the first keypoint falling into the cell, or the average of all the keypoints within the cell. Also, all the keypoints with a single child that are not a start point will be eliminated. An example of the above pruning process is depicted in Figure \ref{fig:keypointsgrid_pruning}. 

We treat the learning problem as combination of segmentation and regression in down-sampled grid space. A lightweight CNN with two 2D convolution layers and 6 residual blocks (Figure \ref{fig:fusionmap_pipeline}) is designed to predict the keypoint grid $K \in \mathbb{R}^{H^\prime \times W^\prime \times 3}$ given the outputs of previous stage. The output keypoint grid essentially encodes the probability of existence of a keypoint in each keypoint grid cell and its relative position within the cell, which is consequently used to refine actual position of keypoints in the original resolution. It is noteworthy that there is no distinction between different types of nodes in this process. 

To train the network, the sum of losses over the pixelwise sigmoid cross entropy to estimate the likelihood of a cell containing a keypoint and mean square error (MSE) of the keypoints coordinates are minimized:
\begin{align}
\begin{split}
  l_{stage2} (R, D, P) = & \lambda_{conf}(-\frac{1}{H^\prime W^\prime}\sum_{i=1}^{H^\prime} \sum_{j=1}^{W^\prime}[p_{ij} \log \hat{p}_{ij} + \\
  & (1-p_{ij}) \log (1-\hat{p}_{ij})]) 
                          + \\ & \lambda_{coord}(\frac{1}{N_k}\sum_{i=1}^{N_k} (c_i-\hat{c}_i)^2) \; 
\end{split}
\end{align}
where $N_k$ represents the predicted total number of keypoints, $p_{ij}^n$ denotes the ground truth map of the $n^{\text{th}}$ keypoint at pixel location $(i, j)$ and $p_{ij}^{\hat{n}}$ is the corresponding sigmoid output at the same location.
\begin{figure*}
     \centering
     \includegraphics[width=0.76\textwidth]{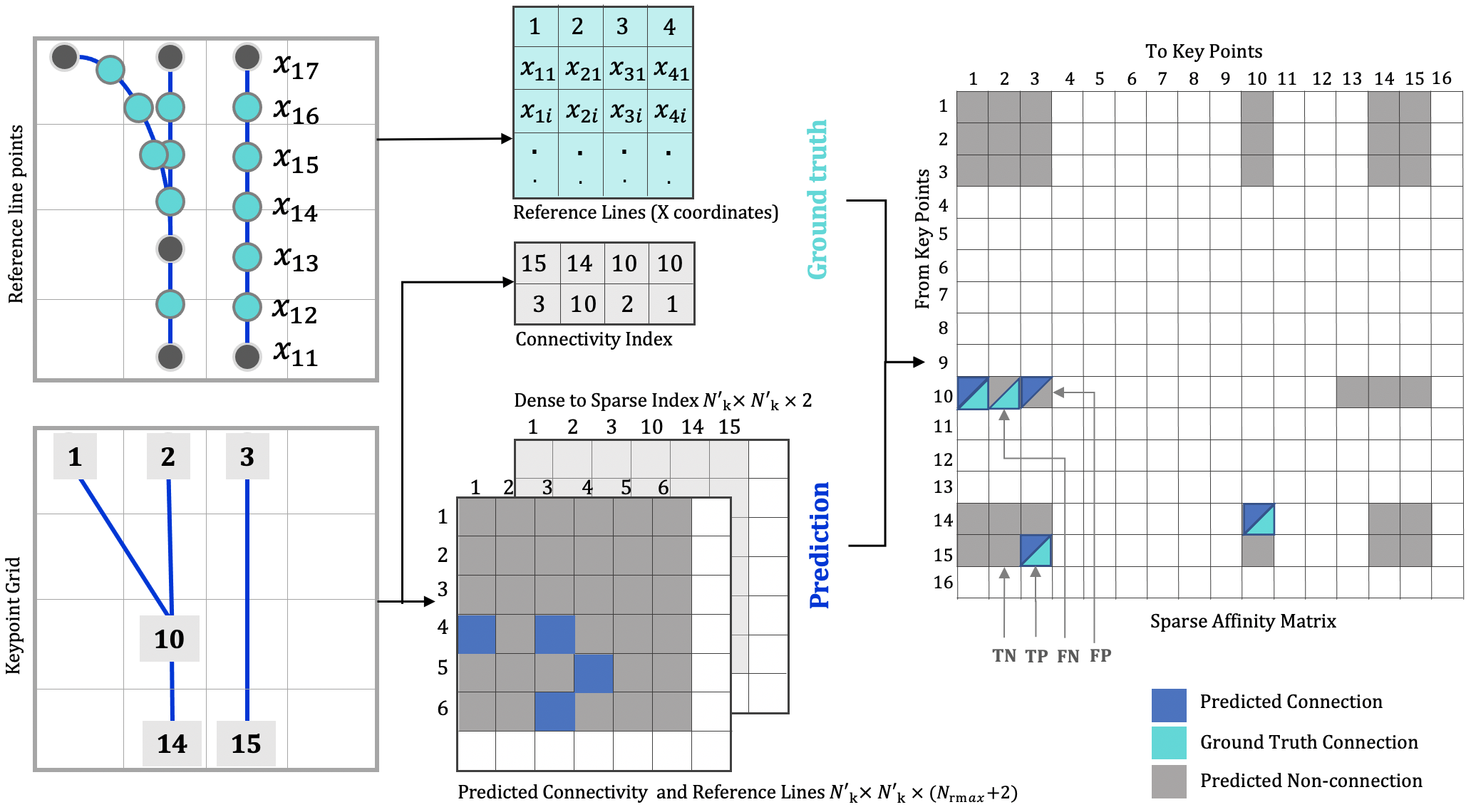}
     \caption{Graph affinity matrix prediction.}
     \label{fig:stage3_process}
\end{figure*}
\subsection{Stage 3: Keypoints Connectivity and Reference Line Prediction}
To complete the graph with the entailing connecting edges to create the road topological structure, the keypoint grid predicted in the previous stage is passed to the third stage of the network to estimate the graph affinity matrix, $p(C, L | K, R, D, P)$, where $C$ and $L$ denote the corresponding connection and lane information of the grid. Hence, in the last stage, in addition to the connections--which is inherently a probability estimation of the existence of a reference line between two keypoints--an accurate localization of the reference line is predicted which is essential to the final task of map-less driving.
\subsubsection{Reference Line Definition} 
For the raw topology generated in Figure \ref{fig:keypointsgrid_pruning}, the definition of a reference line in the HD map might contain a varying number of points depending on the curvature of the underlying line (Figure \ref{fig:refline_points_map}). To simplify the regression task, taking into account the drivable lane orientation constraint explained in \ref{sect:scope_definition}, the reference line is represented as a set of anchor points evenly distributed vertically between two given keypoints. This way, the $y$ coordinates of the reference line points can be calculated and the prediction task is reduced to the regression of $x$ coordinates only. Figure \ref{fig:refline_points_gt} exhibits the above ground truth definition of reference line.
\begin{table*}[t]
\caption{Quantitative evaluation of different stages wrt. grid map resolution.}
\label{table:resolution_analysis}
\centering
\resizebox{0.92\textwidth}{!}{
\begin{tabular}{lcccccccccccccccc}
\toprule
\multicolumn{1}{c}{Resolution} & \multicolumn{2}{c}{\textbf{\shortstack{Distance to\\ref. line}}} & \multicolumn{1}{c}{} & \multicolumn{2}{c}{\textbf{\shortstack{Direction\\of lane}}} & \multicolumn{1}{c}{} & \multicolumn{2}{c}{\textbf{\shortstack{Perp. direction\\of ref. line}}} & \multicolumn{1}{c}{} & \multicolumn{3}{c}{\textbf{\shortstack{Graph\\keypoints}}} & \multicolumn{1}{c}{} & \multicolumn{3}{c}{\textbf{\shortstack{Graph\\connectivity}}}\\
\cmidrule{2-3} \cmidrule{5-6} \cmidrule{8-9} \cmidrule{11-13} \cmidrule{15-17}
& MAE & SSIM & & MAE & SSIM & & MAE & SSIM & & Prec. & Recall & F1 & & Prec. & Recall & F1 \\
\cmidrule[1.5pt]{2-3} \cmidrule[1.5pt]{5-6} \cmidrule[1.5pt]{8-9} \cmidrule[1.5pt]{11-13} \cmidrule[1.5pt]{15-17}
\textbf{128}$\times$\textbf{128} & 0.026 & 0.922 & & 0.016 & 0.931 & & 0.030 & 0.883 & & 0.87 & 0.85 & 0.85 & & 0.71 & 0.97 & 0.79 \\
\textbf{256}$\times$\textbf{256} & 0.015 & 0.956 & & 0.010 & 0.966 & & 0.026 & 0.878 & & 0.80 & 0.71 & 0.75 & & 0.57 & 0.83 & 0.66 \\
\bottomrule
\end{tabular}
}
\end{table*}
\begin{table*}[t]
\caption{Quantitative evaluation of stage 2 and 3 w.r.t. scene complexity, defined based on the total number of keypoints.}
\label{table:difficulty_analysis}
\centering
\resizebox{0.74\textwidth}{!}{
\begin{tabular}{lccccccccc}
\toprule
\multicolumn{1}{l}{Complexity (\# keypoints)} & \multicolumn{1}{c}{} & \multicolumn{3}{c}{\textbf{Graph keypoints}} & \multicolumn{1}{c}{} & \multicolumn{4}{c}{\textbf{Graph connectivity}}\\
\cmidrule{3-5} \cmidrule{7-10}
& & Prec. & Recall & F1 & & Prec. & Recall & F1 & Avg. offset (cm)\\
\cmidrule[1.5pt]{3-5} \cmidrule[1.5pt]{7-10}
\textbf{Easy (1-5)} & & 0.88 & 0.90 & 0.89 & & 0.80 & 0.97 & 0.86 & 15\\
\textbf{Medium (6-10)} & & 0.85 & 0.77 & 0.80 & & 0.61 & 0.95 & 0.69 & 26\\
\textbf{Difficult (11-15)} & & 0.72 & 0.62 & 0.68 & & 0.40 & 0.78 & 0.53 & 42\\
\bottomrule
\end{tabular}
}
\end{table*}
\subsubsection{Graph Affinity Matrix Prediction} 
Theoretically, the keypoint grid can have up to $H^\prime\times W^\prime$ keypoints. In most of the existing road structures, however, empirically this number has a lower limit $N_{k^\prime}$. Hence, rather than having a sparse affinity matrix of size $(H^\prime\times W^\prime)by(H^\prime\times W^\prime)$ to represent all the graph connections, a dense representation of affinity matrix is introduced which keeps the indices of $N_{k^\prime}$ keypoints and their connectivity information. For a given set of keypoints which are represented as a matrix of size $H^\prime\times W^\prime$ labeled with indices $1\dots N_{k^\prime}$, the dense representation would be in the form of two matrices entailing the original connected indices $C\in \mathbb{Z}^{2\times N_{k^\prime}}$ and corresponding reference line information $L\in \mathbb{Z}^{(N_{r_{max}}+2)\times N_{k^\prime}}$, where $N_{r_{max}}$ denotes the maximum number of points to model a reference line segment. For that, the affinity matrix of predictions is initialized with the fully connected set of predicted keypoints. Accordingly, the stage 3 network predicts a dense affinity matrix $N_{k^\prime}\times N_{k^\prime}\times (N_{r_{max}}+2)$ which entails connectivity and reference line information. During the loss calculation process, both ground truth and predicted affinity matrix are mapped back to the sparse matrix. This transformation, also, would prevent accumulation of error from previous stage; i.e. in case of having false predictions, this step ensures that all the correctly predicted keypoints are indexed to the right position in the sparse matrix before calculating the loss. Figure \ref{fig:stage3_process} depicts this process for both stage 3 ground truth and predictions.\\
To construct the road topology and refine the road structure estimated from earlier stages, the following loss function is defined for reference line classification and localization:
\begin{align}
\begin{split}
  l_{stage3} (K, R, D, P) = & \lambda_{conf}(-\frac{1}{N_k^2}\sum_{i=1}^{N_k} \sum_{j=1}^{N_k}[p_{ij} \log \hat{p}_{ij} + \\
  & (1-p_{ij}) \log (1-\hat{p}_{ij})]) \\
                         & + \lambda_{coord}(\frac{1}{N_c N_r}\sum_{i=1}^{N_c}\sum_{j=1}^{N_r} (l_{ij}-\hat{l}_{ij})^2) \;
\end{split}
\end{align}
where $p_{ij}$ denotes the likelihood of existence of connection between keypoints $i$ and $j$, $l$ represents the $x$ coordinates of reference line anchor points, and $N_c$ is the number of connections.
\section{Experiments}
\begin{table*}[t]
\caption{Quantitative evaluation of different stages wrt. input grid map channels. In each experiment the following channels are excluded from input (\ref{sect:data}): (a) Occupancy, (b) Ground semantics, (c) Ground markings, (d) LiDAR intensity, (e) None.}
\label{table:input_analysis}
\centering
\resizebox{0.96\textwidth}{!}{
\begin{tabular}{lccccccccccccccccc}
\toprule
\multicolumn{1}{c}{Input} & \multicolumn{2}{c}{\textbf{\shortstack{Distance to\\ref. line}}} & \multicolumn{1}{c}{} & \multicolumn{2}{c}{\textbf{\shortstack{Direction\\of lane}}} & \multicolumn{1}{c}{} & \multicolumn{2}{c}{\textbf{\shortstack{Perp. direction\\of ref. line}}} & \multicolumn{1}{c}{} & \multicolumn{3}{c}{\textbf{\shortstack{Graph\\keypoints}}} & \multicolumn{1}{c}{} & \multicolumn{4}{c}{\textbf{\shortstack{Graph\\connectivity}}}\\
\cmidrule{2-3} \cmidrule{5-6} \cmidrule{8-9} \cmidrule{11-13} \cmidrule{15-18}
& MAE & SSIM & & MAE & SSIM & & MAE & SSIM & & Prec. & Recall & F1 & & Prec. & Recall & F1 & Avg. Offset(cm) \\
\cmidrule[1.5pt]{2-3} \cmidrule[1.5pt]{5-6} \cmidrule[1.5pt]{8-9} \cmidrule[1.5pt]{11-13} \cmidrule[1.5pt]{15-18 }
\textbf{(a)} & 0.027 & 0.922 & & 0.017 & 0.935 & & 0.031 & 0.885 & & 0.87 & 0.82 & 0.83 & & 0.70 & 0.97 & 0.78 & 27 \\
\textbf{(b)} & 0.025 & 0.928 & & 0.016 & 0.934 & & 0.029 & 0.896 & & 0.89 & 0.80 & 0.84 & & 0.68 & 0.96 & 0.77 & 26 \\
\textbf{(c)} & 0.028 & 0.918 & & 0.018 & 0.935 & & 0.032 & 0.884 & & 0.87 & 0.84 & 0.85 & & 0.69 & 0.97 & 0.77 & 26 \\
\textbf{(d)} & 0.026 & 0.919 & & 0.016 & 0.922 & & 0.034 & 0.879 & & 0.88 & 0.81 & 0.84 & & 0.68 & 0.95 & 0.76 & 28 \\
\textbf{(e)} & 0.026 & 0.922 & & 0.016 & 0.931 & & 0.030 & 0.883 & & 0.87 & 0.85 & 0.85 & & 0.71 & 0.97 & 0.79 & 24 \\
\bottomrule
\end{tabular}
}
\end{table*}
\paragraph{Data} The experiments are done on datasets recorded from Santa Clara county, United States. The data has been collected from multiple passes of several autonomous vehicles equipped with the same set of sensors. The datasets consist of 12,000 frames, with 70 forks/intersections. We leave out one route with 1,000 frames for test of generalization. The recordings are categorized for train and validation with 92:8 ratio.

All ground truth labels (potential ego drivable lanes, directional features, and sparse road topology) are extracted from existing HD maps. Hence, our system has to be trained in areas where HD maps are available, with the goal to generalize to new, previously unmapped areas. The benefit of this approach is that is does not require additional human annotation.

Towards the goal of facilitating further research and baseline comparison, we will release the dataset publicly.
\paragraph{Experimental Setup} For stage 1, the model was trained using Adam with a learning rate of 1e-4, with decay rate and step of 0.96 and 1e+5. In stage 2, the original input to the model was downsized by 8. Finally, in stage 3, for both maximum keypoint and maximum connections value of 16 was chosen based on the distribution of training data. As for the reference point definition, we chose a 4 pixel step. Since all the stages are differentiable, the network is optimized end-to-end to predict the parameters and trained for 200 epochs over the entire dataset.

We use Mean Absolute Error (MAE) and Structural Similarity Index (SSIM) as evaluation metrics for stage 1 tasks. For stage 2 and 3, precision, recall, and F1-score are chosen for evaluation. The average offset from predicted reference line points to ground truth is used to evaluate the performance of points position prediction.
\paragraph{Quantitative Analysis} Due to the lack (to the best of our knowledge) of a public road topology benchmark, comparison with other existing approaches could not be undertaken. Nevertheless, we present our results based on evaluations with our ground truth test data. Table \ref{table:resolution_analysis} presents the detailed evaluation results of each stage for two grid map resolution values of $128\times 128$ and $256\times 256$. As expected, the results of stage 2 and 3 drop by increasing the resolution, since the sensor data gets sparser in the further range and the keypoint prediction has more mis-detections and hence reduced graph connectivity performance. Therefore, for the rest of experiments we only report the values in 128 resolution.

To evaluate the effectiveness of the method towards the ultimate goal of being able to navigate in the areas with no HD map to facilitate self driving at scale, we evaluate the performance of keypoint and reference line point prediction at different scene difficulty levels depending on the complexity of topology in straight roads vs. intersections/forks. As can be seen in Table \ref{table:difficulty_analysis}, even though the increasing number of keypoints affects the performance of system, it is still able to recover the underlying topology.
\paragraph{Ablation Studies} Table \ref{table:input_analysis} outlines the fundamental importance of utilizing different sensor modalities represented as different input channels explained in \ref{sect:data}. By dropping each input channel, we observe a decrease in the model\textquotesingle s performance, among which LiDAR intensity plays the most essential role. Table \ref{table:stage1_analysis} showcases the effect of different stage 1 outputs on the following stages. We can see that excluding direction map or perpendicular direction map leads to much lower performance in stage 3, proving the importance of implicit encoding of direction in the process of learning topology.
\paragraph{Baseline Comparison} To underline the effectiveness of learning connections, we compare our stage 3 network in isolation to a reference method based on the Dijkstra shortest path algorithm. The baseline is described in Algorithm \ref{alg:stage3_baseline}. For the results shown in Table \ref{table:comparison_to_cv_baseline}, ground truth keypoints have been used as input to factor out the influence of stage 2 errors. It can be seen that our learned stage 3 network significantly outperforms the reference method, both in finding correct connections and in reconstructing the correct reference line between two keypoints.
\begin{table}
\caption{Quantitative evaluation of stage 3 compared to the baseline shortest path method.}
\label{table:comparison_to_cv_baseline}
\centering
\resizebox{0.46\textwidth}{!}{
\begin{tabular}{lccc}
\toprule
Method & \multicolumn{3}{c}{\textbf{Graph connectivity}}\\
\cmidrule{2-4}
 & Prec. & Recall & Avg. offset (cm)\\
\cmidrule[1.5pt]{2-4}
\textbf{Proposed method} & 0.996 & 0.994 & 17 \\
\textbf{Shortest path baseline} & 0.78 & 0.78 & 78 \\
\bottomrule
\end{tabular}
}
\end{table}
\begin{algorithm}
\caption{Shortest Path Baseline for Stage 3}
\label{alg:stage3_baseline}
\KwOut{Connectivity $C$; Reference lines $L$}
\KwIn{Distance transform: $R$; Keypoint grid: $K$}
\KwData{Distance threshold: $DT$ (free parameter); Pixel directional steps: $S$ = (L, TL, T, TR, R)}
// Build pixel graph $PG$ in cost image\\
\ForEach{pixel $p1$ in $R$}{
    \ForEach{pixel step $s$ in $S$}{
        pixel $p2$ = $p1$ + $s$\\
        \If{$R(p1) \le DT~\textbf{and}~R(p2) \le DT$}{
            connect $p1$ and $p2$ with weight $R(p2)$\\
        }
    }
}
// Find shortest paths in $PG$ for each keypoint pair\\
Initialize empty connection graph $C$\\
\For{$k1$ in $K$}{
    \For{$k2$ in $K$}{
        shortest path $D$ = Dijkstra($k1, k2, PG$)\\
        \If{$D$ exists}{
            connect $k1$ and $k2$ with weight Dist($D$)\\
            store Path($D$) as ref. line in $L$\\
        }
    }
}
$C$ = MinimumSpanningTree($C$)
\end{algorithm}
\begin{figure*}[t]
     \centering
     \includegraphics[width=0.97\textwidth]{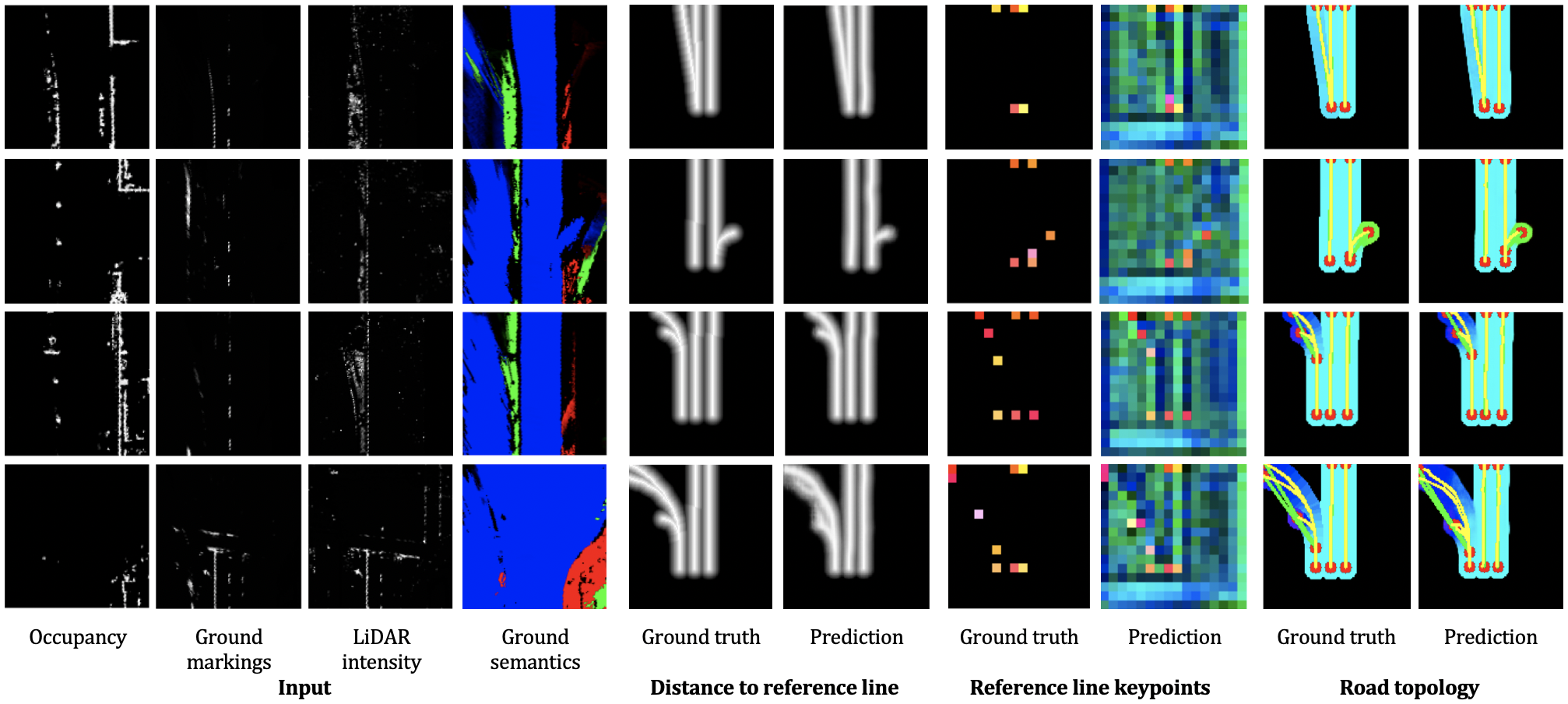}
     \caption{Qualitative results of road topology predictions on the test data.}
     \label{fig:qualitative_results}
\end{figure*}
\begin{figure*}[t]
     \centering
     \includegraphics[width=0.97\textwidth]{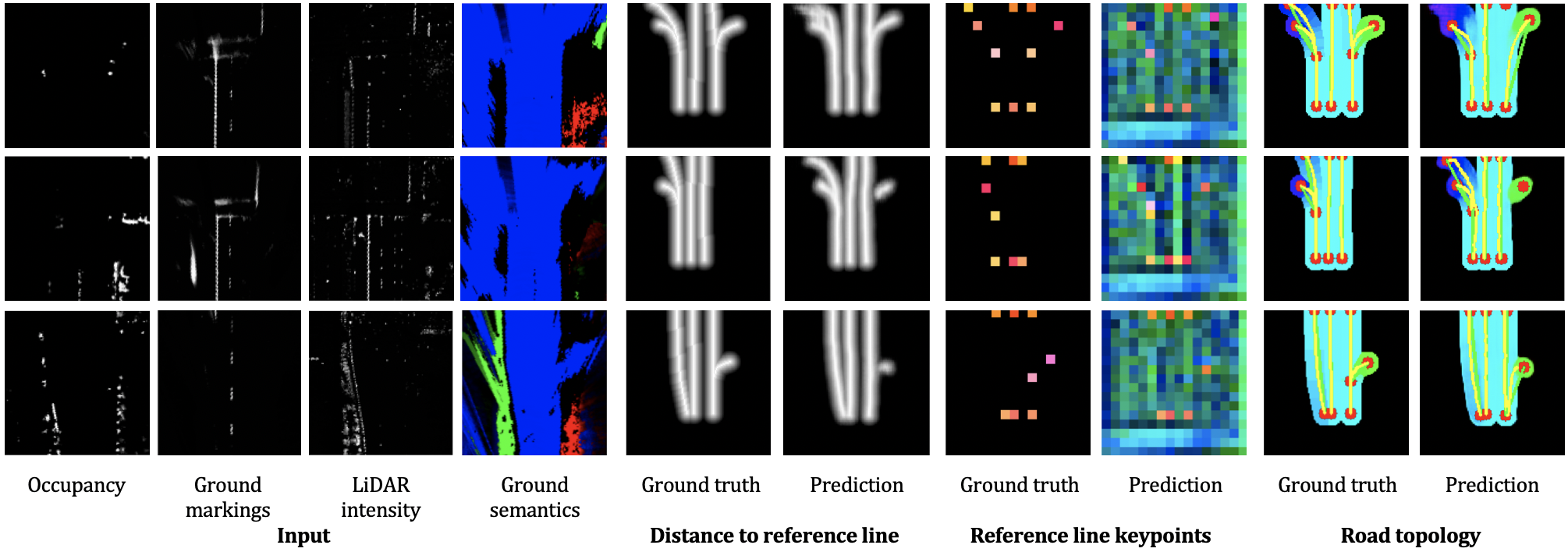}
     \caption{Failure cases.}
     \label{fig:failure_cases}
\end{figure*}
\paragraph{Qualitative Results} Figure \ref{fig:qualitative_results} depicts results obtained from inference of data coming partially from similar areas in training set but driven in the opposite direction with increasing topology complexity. In row 1 and 2, we showcase how our model correctly infers the change of topology by spawning a new lane boundary at a fork. In rows 3 and 4, we demonstrate the behavior of our model at left turn with the option of u-turn in row 4. Figure \ref{fig:failure_cases} displays some failure cases of the system which root in accumulated error of earlier stages or limited sensor information.
\paragraph{Inference Time} For the end-to-end inference of the system the pipeline is optimized to run at 20 Hz. The inference time breakdown for stage 1, 2, and 3 are 7.4, 9.6, and 13.0 (ms), respectively, on a single Tesla P100 GPU. These values are comparable to the experiments with $256\times 256$ grid map resolution taking 9.0, 8.1, and 15.4 (ms).
\begin{table*}
\caption{Quantitative evaluation of the effectiveness of stage 1 outputs. The output subsets are: (a) Distance to ref. line, (b) Distance to ref. line + Direction of lane, (c) Distance to ref. line + Perp. direction of ref. line, and (d) All.}
\label{table:stage1_analysis}
\centering
\resizebox{0.74\textwidth}{!}{
\begin{tabular}{lccccccccc}
\toprule
\multicolumn{1}{l}{Stage 1 output subset} & \multicolumn{1}{c}{} & \multicolumn{3}{c}{\textbf{Graph keypoints}} & \multicolumn{1}{c}{} & \multicolumn{4}{c}{\textbf{Graph connectivity}}\\
\cmidrule{3-5} \cmidrule{7-10}
& & Prec. & Recall & F1 & & Prec. & Recall & F1 & Avg. offset (cm)\\
\cmidrule[1.5pt]{3-5} \cmidrule[1.5pt]{7-10}
\textbf{(a)} & & 0.85 & 0.83 & 0.84 & & 0.66 & 0.88 & 0.74 & 69\\
\textbf{(b)} & & 0.86 & 0.84 & 0.85 & & 0.69 & 0.95 & 0.77 & 45\\
\textbf{(c)} & & 0.85 & 0.84 & 0.85 & & 0.68 & 0.93 & 0.76 & 38\\
\textbf{(d)} & & 0.87 & 0.85 & 0.85 & & 0.71 & 0.97 & 0.79 & 24\\
\bottomrule
\end{tabular}
}
\end{table*}
\section{Conclusion}
In this work, we have presented an approach that directly estimates structured road topology from an autonomous vehicle\textquotesingle s on-board sensors. Our approach copes with various road structures, including one-way streets and large intersections with multiple lanes. Compared to existing approaches, which often require some level of post-processing to improve graph connectivity, fill holes in the prediction through in-painting, need model-based priors or even have a human in-the-loop, our approach estimates the road topology in real time and yields a structure that is directly usable by a behavior planning system, providing an affordable and scalable map solution with fast adaptability and low maintenance. In practice, the predicted map will be utilized by the planning component to  define the best trajectory for the autonomous vehicle based on other agent\textquotesingle s behavior.\\
In the future, to boost the generalization, we plan to invest more into procedural generation of vast amounts of topological scenarios in simulation to address corner cases and create more balanced and diverse datasets. Additionally, we will extend the method to include other semantic map elements, such as stop lines and traffic lights.
\clearpage
\balance
{\small
\bibliographystyle{ieee_fullname}
\bibliography{egbib}

\begin{thebibliography}{10}\itemsep=-1pt

\bibitem{artunedo2019decision}
Antonio Artu{\~n}edo, Jorge Godoy, and Jorge Villagra.
\newblock A decision-making architecture for automated driving without detailed
  prior maps.
\newblock In {\em 2019 IEEE Intelligent Vehicles Symposium (IV)}, pages
  1645--1652. IEEE, 2019.

\bibitem{barnes2017find}
Dan Barnes, Will Maddern, and Ingmar Posner.
\newblock Find your own way: Weakly-supervised segmentation of path proposals
  for urban autonomy.
\newblock In {\em 2017 IEEE International Conference on Robotics and Automation
  (ICRA)}, pages 203--210. IEEE, 2017.

\bibitem{bastani2018roadtracer}
Favyen Bastani, Songtao He, Sofiane Abbar, Mohammad Alizadeh, Hari
  Balakrishnan, Sanjay Chawla, Sam Madden, and David DeWitt.
\newblock Roadtracer: Automatic extraction of road networks from aerial images.
\newblock In {\em Proceedings of the IEEE Conference on Computer Vision and
  Pattern Recognition}, pages 4720--4728, 2018.

\bibitem{beck2014non}
Johannes Beck and Christoph Stiller.
\newblock Non-parametric lane estimation in urban environments.
\newblock In {\em 2014 IEEE Intelligent Vehicles Symposium Proceedings}, pages
  43--48. IEEE, 2014.

\bibitem{chaurasia2017linknet}
Abhishek Chaurasia and Eugenio Culurciello.
\newblock Linknet: Exploiting encoder representations for efficient semantic
  segmentation.
\newblock In {\em 2017 IEEE Visual Communications and Image Processing (VCIP)},
  pages 1--4. IEEE, 2017.

\bibitem{chiu2005lane}
Kuo-Yu Chiu and Sheng-Fuu Lin.
\newblock Lane detection using color-based segmentation.
\newblock In {\em IEEE Proceedings. Intelligent Vehicles Symposium, 2005.},
  pages 706--711. IEEE, 2005.

\bibitem{Dempster1968}
A.~P. Dempster.
\newblock A generalization of bayesian inference.
\newblock {\em Journal of the Royal Statistical Society. Series B
  (Methodological)}, 30(2):205--247, 1968.

\bibitem{dickmanns1992recursive}
Ernst~D. Dickmanns and Birger~D. Mysliwetz.
\newblock Recursive 3-d road and relative ego-state recognition.
\newblock {\em IEEE Transactions on Pattern Analysis \& Machine Intelligence},
  (2):199--213, 1992.

\bibitem{he2016accurate}
Bei He, Rui Ai, Yang Yan, and Xianpeng Lang.
\newblock Accurate and robust lane detection based on dual-view convolutional
  neutral network.
\newblock In {\em 2016 IEEE Intelligent Vehicles Symposium (IV)}, pages
  1041--1046. IEEE, 2016.

\bibitem{HernandezJuarez2019SlantedSA}
Daniel Hernandez-Juarez, Lukas Schneider, Pau Cebrian, Antonio Espinosa, David
  V{\'a}zquez, Antonio~M. L{\'o}pez, Uwe Franke, Marc Pollefeys, and Juan~C.
  Moure.
\newblock Slanted stixels: A way to represent steep streets.
\newblock {\em International Journal of Computer Vision}, 127:1643 -- 1658,
  2019.

\bibitem{homayounfar2018hierarchical}
Namdar Homayounfar, Wei-Chiu Ma, Shrinidhi Kowshika~Lakshmikanth, and Raquel
  Urtasun.
\newblock Hierarchical recurrent attention networks for structured online maps.
\newblock In {\em Proceedings of the IEEE Conference on Computer Vision and
  Pattern Recognition}, pages 3417--3426, 2018.

\bibitem{homayounfar2019dagmapper}
Namdar Homayounfar, Wei-Chiu Ma, Justin Liang, Xinyu Wu, Jack Fan, and Raquel
  Urtasun.
\newblock Dagmapper: Learning to map by discovering lane topology.
\newblock In {\em Proceedings of the IEEE International Conference on Computer
  Vision}, pages 2911--2920, 2019.

\bibitem{kong2010general}
Hui Kong, Jean-Yves Audibert, and Jean Ponce.
\newblock General road detection from a single image.
\newblock {\em IEEE Transactions on Image Processing}, 19(8):2211--2220, 2010.

\bibitem{kuhnl2012spatial}
Tobias K{\"u}hnl, Franz Kummert, and Jannik Fritsch.
\newblock Spatial ray features for real-time ego-lane extraction.
\newblock In {\em 2012 15th International IEEE Conference on Intelligent
  Transportation Systems}, pages 288--293. IEEE, 2012.

\bibitem{lee2017vpgnet}
Seokju Lee, Junsik Kim, Jae Shin~Yoon, Seunghak Shin, Oleksandr Bailo, Namil
  Kim, Tae-Hee Lee, Hyun Seok~Hong, Seung-Hoon Han, and In So~Kweon.
\newblock Vpgnet: Vanishing point guided network for lane and road marking
  detection and recognition.
\newblock In {\em Proceedings of the IEEE international conference on computer
  vision}, pages 1947--1955, 2017.

\bibitem{lee2018development}
Unghui Lee, Jiwon Jung, Seokwoo Jung, and David~Hyunchul Shim.
\newblock Development of a self-driving car that can handle the adverse
  weather.
\newblock {\em International journal of automotive technology}, 19(1):191--197,
  2018.

\bibitem{li2016deep}
Jun Li, Xue Mei, Danil Prokhorov, and Dacheng Tao.
\newblock Deep neural network for structural prediction and lane detection in
  traffic scene.
\newblock {\em IEEE transactions on neural networks and learning systems},
  28(3):690--703, 2016.

\bibitem{liang2019convolutional}
Justin Liang, Namdar Homayounfar, Wei-Chiu Ma, Shenlong Wang, and Raquel
  Urtasun.
\newblock Convolutional recurrent network for road boundary extraction.
\newblock In {\em Proceedings of the IEEE Conference on Computer Vision and
  Pattern Recognition}, pages 9512--9521, 2019.

\bibitem{marmanis2016semantic}
Dimitrios Marmanis, Jan~D Wegner, Silvano Galliani, Konrad Schindler, Mihai
  Datcu, and Uwe Stilla.
\newblock Semantic segmentation of aerial images with an ensemble of cnss.
\newblock {\em ISPRS Annals of the Photogrammetry, Remote Sensing and Spatial
  Information Sciences, 2016}, 3:473--480, 2016.

\bibitem{mattyus2017deeproadmapper}
Gell{\'e}rt M{\'a}ttyus, Wenjie Luo, and Raquel Urtasun.
\newblock Deeproadmapper: Extracting road topology from aerial images.
\newblock In {\em Proceedings of the IEEE International Conference on Computer
  Vision}, pages 3438--3446, 2017.

\bibitem{meyer2018deep}
Annika Meyer, N~Ole Salscheider, Piotr~F Orzechowski, and Christoph Stiller.
\newblock Deep semantic lane segmentation for mapless driving.
\newblock In {\em 2018 IEEE/RSJ International Conference on Intelligent Robots
  and Systems (IROS)}, pages 869--875. IEEE, 2018.

\bibitem{mnih2010learning}
Volodymyr Mnih and Geoffrey~E Hinton.
\newblock Learning to detect roads in high-resolution aerial images.
\newblock In {\em European Conference on Computer Vision}, pages 210--223.
  Springer, 2010.

\bibitem{neven2018towards}
Davy Neven, Bert De~Brabandere, Stamatios Georgoulis, Marc Proesmans, and Luc
  Van~Gool.
\newblock Towards end-to-end lane detection: an instance segmentation approach.
\newblock In {\em 2018 IEEE intelligent vehicles symposium (IV)}, pages
  286--291. IEEE, 2018.

\bibitem{ort2019maplite}
Teddy Ort, Krishna Jatavallabhula, Rohan Banerjee, Sai~Krishna Gottipati,
  Dhaivat Bhatt, Igor Gilitschenski, Liam Paull, and Daniela Rus.
\newblock Maplite: Autonomous intersection navigation without a detailed prior
  map.
\newblock {\em IEEE Robotics and Automation Letters}, 2019.

\bibitem{piewak2018improved}
Florian Piewak, Peter Pinggera, Markus Enzweiler, David Pfeiffer, and Marius
  Z{\"o}llner.
\newblock Improved semantic stixels via multimodal sensor fusion.
\newblock In {\em German Conference on Pattern Recognition}, pages 447--458.
  Springer, 2018.

\bibitem{suleymanov2018inferring}
Tarlan Suleymanov, Paul Amayo, and Paul Newman.
\newblock Inferring road boundaries through and despite traffic.
\newblock In {\em 2018 21st International Conference on Intelligent
  Transportation Systems (ITSC)}, pages 409--416. IEEE, 2018.

\bibitem{topfer2014efficient}
Daniel T{\"o}pfer, Jens Spehr, Jan Effertz, and Christoph Stiller.
\newblock Efficient road scene understanding for intelligent vehicles using
  compositional hierarchical models.
\newblock {\em IEEE Transactions on Intelligent Transportation Systems},
  16(1):441--451, 2014.

\end{thebibliography}
}

\end{document}